\documentclass[pdflatex,sn-mathphys,Numbered]{sn-jnl}

\usepackage{graphicx}%
\usepackage{multirow}%
\usepackage{amsmath,amssymb,amsfonts}%
\usepackage{mathrsfs}%
\usepackage[title]{appendix}%
\usepackage{xcolor}%
\usepackage{textcomp}%
\usepackage{manyfoot}%
\usepackage{booktabs}%
\usepackage{algorithm}%
\usepackage{algorithmicx}%
\usepackage{algpseudocode}%
\usepackage{listings}%

\begin{document}

\title[HyPE: Attention with Hyperbolic Biases for Relative Positional Encoding]{HyPE: Attention with Hyperbolic Biases for Relative Positional Encoding}


\author{\fnm{Giorgio} \sur{Angelotti}\footnote{Dr. Giorgio Angelotti, a physicist, data scientist, and machine learning engineer, earned his PhD in Computer Science from the University of Toulouse, France, in 2023.}}\email{giorgio@thegiorgio.org}

\affil{Independent Researcher}


\abstract{In Transformer-based architectures, unmasked self-attention without positional information is permutation-equivariant with respect to the input sequence's tokens. To impose sequential order, token positions are typically encoded using a scheme with either fixed or learnable parameters. We introduce Hyperbolic Positional Encoding (HyPE), a method that utilizes hyperbolic functions' properties to encode tokens' relative positions. This approach biases the attention mechanism without the necessity of storing the $O(L^2)$ values of the bias tensor, with $L$ being the length of the input sequence. HyPE leverages preliminary concatenation operations and matrix multiplications, encoding relative distances by indirectly incorporating biases into the softmax computation. This design supports gradient backpropagation through the encoding and can be used with FlashAttention-2 under the width and scaling conditions below. We analytically demonstrate that, by careful hyperparameter selection, HyPE can approximate the causal attention bias of ALiBi \citep{press2021train} on a bounded range; model generalization beyond the pretraining length remains to be evaluated. The experimental evaluation of HyPE is proposed as a direction for future research.}


\maketitle

{\small\noindent\textbf{Revision note (October 2026).} This revision corrects sign, indexing and query/key notation, including the causal ALiBi parameter mapping. It retains the two-coordinate hyperbolic factorization and states the ALiBi approximation on a bounded range. Attention scaling, padding, memory complexity and historical FlashAttention support are clarified. No model-training or length-extrapolation results are added.\par}

\section{Introduction}
Transformer-based networks \citep{vaswani2017attention} have become pervasive across various fields, including Natural Language Processing \citep{devlin2018bert, brown2020language, almazrouei2023falcon, touvron2023llama}, Computer Vision \citep{dosovitskiy2021an, oquab2023dinov2}, and Offline Reinforcement Learning \citep{chen2021decision,janner2021offline}. Within Transformer-based architectures, the attention mechanism in the Transformer block hones in on the interrelations between token pairs in the input sequence. However, without positional information or a position-dependent mask, this attention mechanism is permutation-equivariant, which implies that Transformers do not inherently discern the absolute and relative positions of tokens in a sequence. To remedy this, Positional Encoding (employing either fixed or learnable parameters) is integrated into the architecture \citep{vaswani2017attention,shaw2018self}, thereby augmenting the effectiveness of Transformers. Nevertheless, these encoding strategies introduce limitations to their versatility, particularly in their ability to adapt. Specifically, learnable absolute positional encodings, tailored to the length of the input sequence, become fixed upon training completion, in parallel with the network weights. This inflexibility poses significant challenges when dealing with sequences at testing time that surpass the lengths encountered during training \citep{press2021train}.

Subsequent sections will delve into the integration of positional encoding in Transformer-based architectures, highlighting the existing limitations and providing the rationale for the innovations presented in this study.

\section{Positional Encoding}
Let us first recap what is computed by the attention mechanism.

Let $X \in \mathbb{R}^{L\times d}$ represent the sequence of embedded tokens, with $Q, K, V \in \mathbb{R}^{L\times d}$ denoting the query, key, and value matrices, respectively, and $W^{Q}, W^{K}, W^{V} \in \mathbb{R}^{d \times d}$ being the corresponding linear operators. We take $L$ and $d$ to be positive integers and apply softmax row-wise over a nonempty set of allowed keys. The attention mechanism's output (for a single head) is then calculated as:
\begin{equation}
\label{eq:attention}
O = \textrm{softmax}\left(\frac{QK^{T}}{\sqrt{d}}\right)V.
\end{equation}
\citet{vaswani2017attention} introduced positional information into Transformer models by superimposing a sinusoidal signal (dependent on the token's absolute position) onto the $d$-dimensional embedding resulting from the initial embedding layer. This sinusoidal bias represents the token's position within the sequence through a phase that varies with its position, with frequencies varying across embedding coordinates. This bias $a^{\textrm{Sin}} \in \mathbb{R}^{L \times d}$ is usually directly added to the embedding before computing the queries, keys and values: $X \leftarrow X + a^{\textrm{Sin}}$.

In pursuit of better generalization for larger contexts and accommodating sequences beyond pre-training lengths, it is argued that \textit{relative distances} should be prioritized over \textit{absolute positions} within the encoded representation. To this end, \citet{shaw2018self} proposed Relative Positional Representations. The work in \citep{shaw2018self} modifies keys and values by summing to them learnable embeddings, whose parameters are adjusted based on the relative positions of tokens. Following a similar approach, \citet{huang2020improve} expanded this method to absolute distance encoding and vector embeddings. These approaches, in spite of their performances, can be computationally expensive. In the T5 model \citep{raffel2020exploring}, an attention bias $a^{\textrm{T5}} \in \mathbb{R}^{L\times L}$ (a function of the relative distances) is integrated into the attention computation to encode the relative positions.
Dense implementations of such relative positional encodings store an $O(L^2)$ bias tensor, but this storage is not required by all implementations. Moreover, recent algorithmic advances enabling $O(Ld)$ memory usage for self-attention retain quadratic arithmetic and do not all support bias addition before the softmax \citep{dao2022flashattention, dao2023flashattention}. As of 30 October 2023, the cited experimental Triton implementation does not support gradients for attention bias parameters in FlashAttention-1\footnote{Reference to the official Triton implementation: \url{https://github.com/Dao-AILab/flash-attention/blob/83aef842beec1037eb8c1d9c3ef3ed8aae80b091/flash_attn/flash_attn_triton.py}} \citep{dao2022flashattention}, while FlashAttention-2\footnote{Reference to the official implementation: \url{https://github.com/Dao-AILab/flash-attention/blob/83aef842beec1037eb8c1d9c3ef3ed8aae80b091/flash_attn/flash_attn_interface.py}}\citep{dao2023flashattention} does not yet support an additive bias input.\footnote{These statements concern the cited 2023 implementations. Native ALiBi support was added in v2.4.0 (22 December 2023): \url{https://github.com/Dao-AILab/flash-attention/releases/tag/v2.4.0}.}

Rotary Positional Encoding (RoPE)\citep{su2021roformer} includes both absolute positions and relative distances. RoPE uses a block diagonal rotation matrix to rotate each token's projected query and key. This encodes the position by rotating coordinate pairs within each query and key vector. The varying angle of rotation with sequence position ensures the encoding of absolute positions, while the relative position is captured through the product of the query and key rotations.
Since RoPE is injected through matmul operations and does not leverage learned parameters, it avoids a separate positional bias tensor; runtime comparisons with T5 depend on the implementation.

Nevertheless, generalization to sequences of length $L_{\textrm{Extra}} > L$, where $L$ is the pre-training length, is not guaranteed \citep{press2021train}. To tackle this issue, the Attention with Linear Biases (ALiBi) paradigm \citep{press2021train} introduces a static, non-learned linear relative positional bias directly into the softmax argument, efficiently scaling Transformer architectures for extrapolation on longer than training context windows. However, a dense ALiBi implementation stores an $O(L^2)$ bias tensor, although the bias formula does not require this storage. Its support in the cited 2023 implementations is as described above.

Our research seeks to answer:

\noindent\emph{Can we devise a nonlinear relative positional bias that incorporates learnable parameters and can be used with FlashAttention-2?}

We introduce Hyperbolic Positional Encoding (HyPE), a differentiable nonlinear relative-position bias with a two-coordinate factorization. Instead of adding a relative distance bias to the softmax term, HyPE utilizes hyperbolic functions and matrix multiplication properties to incorporate the information into the $QK^{T}$ computation. This method can be used with FlashAttention under the conditions below, eliminating the need to store the full $O(L^2)$ bias tensor by \textit{indirectly} calculating it during matrix multiplication.

The forthcoming section provides a concise overview of hyperbolic functions and our principal contribution.
\vspace{-0.1cm}
\section{Hyperbolic Relative Positional Encoding}
Let us examine the hyperbolic sine function:
\begin{equation}
\label{eq:sinh}
2\sinh{(x)} = e^{x}-e^{-x},
\end{equation}
which can be expressed as a truncated series
\begin{equation}
\label{eq:approx}
\sinh{(x)} = x + O(x^{3}) \quad \text{as} \quad x \to 0.
\end{equation}
It follows from Equation \ref{eq:sinh} that
\begin{equation}
\label{eq:sum}
2\sinh{(x-y)} = e^{x-y} - e^{y-x}.
\end{equation}
This trivial identity will be fundamental for the ensuing discussion.

Define $a ^ {\textrm{HyPE}} \in \mathbb{R}^{L \times L}$ such that
\begin{equation}
a ^ {\textrm{HyPE}}_{i,j} = -\tau \sinh{(\mu(j-i))}.
\end{equation}
Our goal is to efficiently integrate the hyperbolic positional bias $a^{\textrm{HyPE}}$ into the softmax function's argument in Equation \ref{eq:attention}.

Let $\eta^{\{Q,K\}} \in \mathbb{R}^{L\times 2}$ be:
\begin{equation}
\eta^{Q}_{i,j} = \dfrac{\tau\sqrt{d}}{2}e^{(-1)^{j+1}\mu i}
\end{equation}
\begin{equation}
\eta^{K}_{i,j} = (-1)^{j+1}e^{(-1)^{j}\mu i}
\end{equation}
with token index $i \in \{0,\dots,L-1\}$ and feature index $j \in \{0,1\}$ and parameters $(\mu,\tau) \in \mathbb{R}^2$. Here, $\mu$ sets the distance scale and $\tau$ the signed amplitude; the slope at zero displacement is $-\tau\mu$.

\noindent By concatenating $\eta^{\{Q,K\}}$ with $Q$ and $K$ respectively:
\begin{equation}
\label{eq:concat1}
\hat{Q} = \textrm{concat}(Q, \eta^{Q}),
\end{equation}
\begin{equation}
\label{eq:concat2}
\hat{K} = \textrm{concat}(K, \eta^{K}),
\end{equation}
we can express the product $\hat{Q}\hat{K}^{T}$ as:
\begin{equation}
\hat{Q}\hat{K}^{T} = QK^{T} + \sqrt{d} a ^ {\textrm{HyPE}}.
\end{equation}
To illustrate, consider the explicit matrix multiplication:
\begin{equation}
\begin{aligned}
(\hat{Q}\hat{K}^{T})_{i,l}
 &= \sum_{j=0}^{d+1}\hat{Q}_{i,j}\hat{K}_{l,j} \\
 &= \sum_{j=0}^{d-1}\hat{Q}_{i,j}\hat{K}_{l,j}
    + \sum_{j=d}^{d+1}\hat{Q}_{i,j}\hat{K}_{l,j} \\
 &= \sum_{j=0}^{d-1}Q_{i,j}K_{l,j}
    + \sum_{j=0}^{1}\eta^{Q}_{i,j}\eta^{K}_{l,j}.
\end{aligned}
\end{equation}
By utilizing the earlier identity (Equation \ref{eq:sum}):
\begin{equation}
\begin{aligned}
\sum_{j=0}^{1}\eta^{Q}_{i,j}\eta^{K}_{l,j}
 &= \dfrac{\tau \sqrt{d}}{2}\left(-e^{\mu(l-i)}+e^{\mu(i-l)}\right) \\
 &= -\tau \sqrt{d}\sinh{(\mu(l-i))}
  = \sqrt{d}\,a^{\textrm{HyPE}}_{i,l}.
\end{aligned}
\end{equation}
Subsequently, the attention output can be written:
\begin{equation}
\hat{O} = \textrm{softmax} \left( \dfrac{\hat{Q}\hat{K}^{T}}{\sqrt{d}} \right)V = \textrm{softmax} \left( \dfrac{QK^{T}}{\sqrt{d}} + a^{\textrm{HyPE}}\right)V.
\end{equation}

\noindent In this fashion, the $a^{\textrm{HyPE}}$ bias is incorporated without explicit storage, calculated indirectly during matrix multiplication.

To meet tensor-width requirements, $\eta^{\{Q,K\}}$ can be stacked $n\geq1$ times, resulting in a dimension of $d+2n$. This multiplies the bias by $n$; divide $\tau$ by $n$ in each copy to preserve it. Alternatively, pad with zeros. Keep the original $1/\sqrt{d}$ scale and the causal mask when required. Implementations requiring equal query, key and value widths must also pad $V$ with zeros and discard its added output coordinates. Head-width limits still apply.

\subsection{Approximate Equivalence with ALiBi}
Using the truncated series expression (Equation \ref{eq:approx}), for causal query $i$ and key $j\leq i$, set $\mu>0$ and $\tau=-m/\mu$, where $m>0$ is ALiBi's slope and $a^{\textrm{ALiBi}}_{i,j}=-m(i-j)$. Let $D\geq0$ be the maximum query--key separation considered, so $|j-i|\leq D$. We obtain
\begin{equation}
\begin{aligned}
a^{\textrm{HyPE}}(\mu,-m/\mu)_{i,j}
 &= a^{\textrm{ALiBi}}(m)_{i,j}+r_{i,j}, \\
|r_{i,j}| &\leq \frac{m\mu^2|j-i|^3}{6}\cosh(\mu D).
\end{aligned}
\end{equation}
This result suggests that for an accurate selection of (hyper)parameters HyPE can approximate ALiBi's causal bias, while using FlashAttention-2 under the conditions above. For comparison, causal ALiBi's attention weights have an exact one-coordinate realization in real arithmetic: $q'_i=[q_i,1]$ and $k'_j=[k_j,m\sqrt{d}\,j]$. At the original $1/\sqrt{d}$ scale, the added $mj$ term differs from $m(j-i)$ by a row constant, leaving softmax unchanged on the same causal support. This construction also permits learning a positive head-specific slope $m$, subject to the same implementation conditions. HyPE provides a nonlinear hyperbolic bias family through two appended coordinates.
\subsection{Selection of (hyper)parameters}
Given the exponential divergence of $\sinh(x)$ for large $|x|$, we suggest choosing $\mu>0$ so that Equation (14), evaluated at $|j-i|=D=L_{\textrm{Extra}}-1$, meets the desired error tolerance, where $L_{\textrm{Extra}}$ exceeds the pre-training length $L$.

The amplitude $\tau$ may be static or dynamically adjusted through gradient descent. Learned values need not satisfy the approximation conditions.
\subsection{Multi-Head Attention Implementation}
HyPE accommodates multi-head attention by defining head-specific $\eta_{h}^{\{Q,K\}}$ and applying the concatenation steps. In this multi-head context, each head can have a unique distance-scale parameter $\mu_{h}$, and $\tau_{h}$ could be a learnable amplitude.

\subsection{Extending HyPE to Images and Multidimensional Data}
HyPE is versatile and can be extended to process multidimensional data, such as images. For instance, consider a 3D image structured as a parallelepiped with dimensions $(L_{x}, L_{y}, L_{z})$. Here, each voxel represents a token, identifiable by its 3D coordinates $(i,j,k)$.
These voxels are fed into the Transformer model as a flattened sequence of tokens with a total count of $L_{x}L_{y}L_{z}$.

To adapt HyPE for this context, one must carefully align the $\eta_{{x,y,z}}^{\{Q,K\}}$ matrices, ensuring that each token's dimensional indices are accurately represented.   It is crucial to verify that the query and key feature pairs for the same axis interact during matrix multiplication, using two coordinates per axis in matching axis order. This precise arrangement allows the computation of attention as follows:
\begin{equation}
\hat{O} = \textrm{softmax} \left(\dfrac{QK^{T}}{\sqrt{d}} + a^{\textrm{HyPE}}_{x} + a^{\textrm{HyPE}}_{y} + a^{\textrm{HyPE}}_{z}\right)V.
\end{equation}
The same reasoning can be extended to input of any dimension, giving a sum of signed per-axis biases.

\section{Experimental Evaluation}

The author, as an Independent Researcher with limited access to computational resources, has deferred the experimental evaluation of the algorithm to future work. We posit that a partial evaluation, constrained by computational limitations, might lead to an incomplete analysis that might not fully represent the merits of the proposed approach. Nonetheless, we invite researchers worldwide to apply the technique within their own computational environments and contribute to its empirical evaluation.
The primary challenges in implementing this approach may lie in rounding, cancellation and overflow in low-precision values and the allocation of memory. Equation (14) is a real-arithmetic bound. In the exponential-feature implementation, reducing $\mu$ at fixed $m$ with $\tau=-m/\mu$ can increase feature magnitudes and cancellation; floating-point output and gradient accuracy require separate evaluation.
\section{Conclusion}

This study tackled the challenge of devising a differentiable nonlinear Relative Positional Encoding while leveraging the advancements in FlashAttention technology. Our development of Hyperbolic Positional Encoding (HyPE) requires storing $4Lh$ positional values in one dimension (where $h$ represents the number of attention heads) and is capable of implicitly integrating a hyperbolic attention bias into the softmax argument of the attention mechanism via matrix multiplication of altered queries and keys. This count excludes the expanded activations. Materializing the concatenations necessitates additional memory for queries and keys, and padded values when needed, before attention is computed using a compatible FlashAttention implementation with the original scale. While the parameters of HyPE have the potential to be learnable, we demonstrate that with the sign and range conditions above, it can approximate the causal attention bias of ALiBi. We also show that HyPE can incorporate relative positions for input of any dimension, e.g. images. Comprehensive experimental evaluation remains an objective for future work.
\backmatter

\setlength{\bibsep}{0.75em}
\bibliography{sn-bibliography}
\end{document}